\documentclass[letterpaper, 10 pt, conference]{ieeeconf}  % Comment this line out if you need a4paper

\IEEEoverridecommandlockouts                              % This command is only needed if 
\usepackage[utf8]{inputenc}
\usepackage{graphicx}
\usepackage{amsmath, amssymb}
\let\labelindent\relax
\usepackage{enumitem}
\usepackage{siunitx}

\usepackage{booktabs}
\usepackage{placeins} % for \FloatBarrier to control float placement
\usepackage{cite}
\usepackage[hidelinks]{hyperref}

\setlist{itemsep=0pt, topsep=2pt, parsep=0pt, partopsep=0pt}

\title{\LARGE \bf
A Deployment Case Study in Robotic Apparel Automation: Digital Twin Integration, Interoperability, and Workforce Enablement}

\begin{document}

\author{Gokul Narayanan$^{1}$, Abhiroop Ajith$^{1}$, Jonathan Zornow$^{2}$, Carlos Calle$^{3}$, Auralis Herrero Lugo$^{4}$, \\ Jose Luis Susa Rincon$^{1}$,   Chengtao Wen$^{1}$ and Eugen Solowjow$^{1}$% <-this % stops a space
% \thanks{$^{*}$ These authors contributed equally to this work.}
\thanks{$^{1}$Abhiroop Ajith, Gokul Narayanan, Jose Luis Susa Rincon, Eugen Solowjow, Chengtao Wen are with Siemens Corporation
        {\tt\small \{abhiroop.ajith, gokul.sathya\_narayanan,
        eugen.solowjow,chengtao.wen\}@siemens.com}}%
\thanks{$^{2}$ Jonathan Zornow is with Sewbo.
    {\tt\small \{jon@sewbo.com\}}}%
\thanks{$^{3}$Carlos Calle is with Levi's. 
    {\tt\small \{ccalle@levi.com\}}}%
\thanks{$^{4}$Auralis Herrero Lugo is with Bluewater Defense.
        {\tt\small \{aherrero@bwdefense.com\}}}%
}   

\maketitle
\thispagestyle{empty}
\pagestyle{empty}

%%%%%%%%%%%%%%%%%%%%%%%%%%%%%%%%%%%%%%%%%%%%%%%%%%%%%%%%%%%%%%%%%%%%%%%%%%%%%%%%

\begin{abstract}
Despite steady advances in flexible automation in sectors such as electronics and automotive manufacturing, apparel automation remains challenging because fabrics are deformable and difficult to manipulate with robots.
% shop floors often rely on legacy and heterogeneous equipment, and high-throughput production leaves little tolerance for variability, downtime, or rework. 
This paper presents a deployment-oriented case study of a robotic sewing system for denim manufacturing, emphasizing the system-level integration required for practical adoption. At the engineering level, a digital thread module parses DXF production drawings into process parameters and executable robot trajectories, reducing manual programming effort and enabling rapid re-targeting across sewing operations. In parallel, a digital twin of the workcell is used during pre-deployment to validate reach and clearance, refine layout and sequencing, evaluate operator access, and assess cycle-time compatibility with upstream and downstream tasks, thereby reducing commissioning risk. At deployment, the system integrates a collaborative robot with conventional sewing equipment, welding, suction fixtures, and machine-level controllers through an interoperability layer. Runtime monitoring and verification, including seam monitoring, collision checking, and trajectory-level validation, improve robustness under environmental variability, while operator-facing training and guidance tools support setup, troubleshooting, and technology adoption. 
Two staged factory deployments on denim shorts, covering 2D pocket operations and 3D garment-shaping seams, show that digital-twin-based validation, digital-thread-driven task generation, interoperability, runtime verification, and operator training are important for scaling robotic apparel automation. An illustrative  video of the second-stage deployment is available at \url{https://youtu.be/jL-mlLc7_Fg}.

\end{abstract}

%%%%%%%%%%%%%%%%%%%%%%%%%%%%%%%%%%%%%%%%%%%%%%%%%%%%%%%%%%%%%%%%%%%%%%%%%%%%%%%%

\section{Introduction}

Apparel manufacturing remains one of the least automated industrial sectors because of the challenges in handling fabrics. 
As a result, apparel production still relies heavily on manual labor and semi-automated machines designed for isolated operations, limiting flexibility and reconfigurability~\cite{Schlegl2018RobotsIT,Nguyen2022,Jindal2021}.
Prior work has advanced automated sewing through hierarchical fabric control \cite{KOUSTOUMPARDIS201434}, end-to-end sewing pipelines for specialized garments \cite{pr9020289}, multi-robot sewing systems \cite{6224880,Schrimpft2015}, open-loop fabric control \cite{Winck2009}, and real-time seam correction \cite{KOSAKA2023103005}.

\begin{figure}[t]
  \centering
  \includegraphics[width=\linewidth]{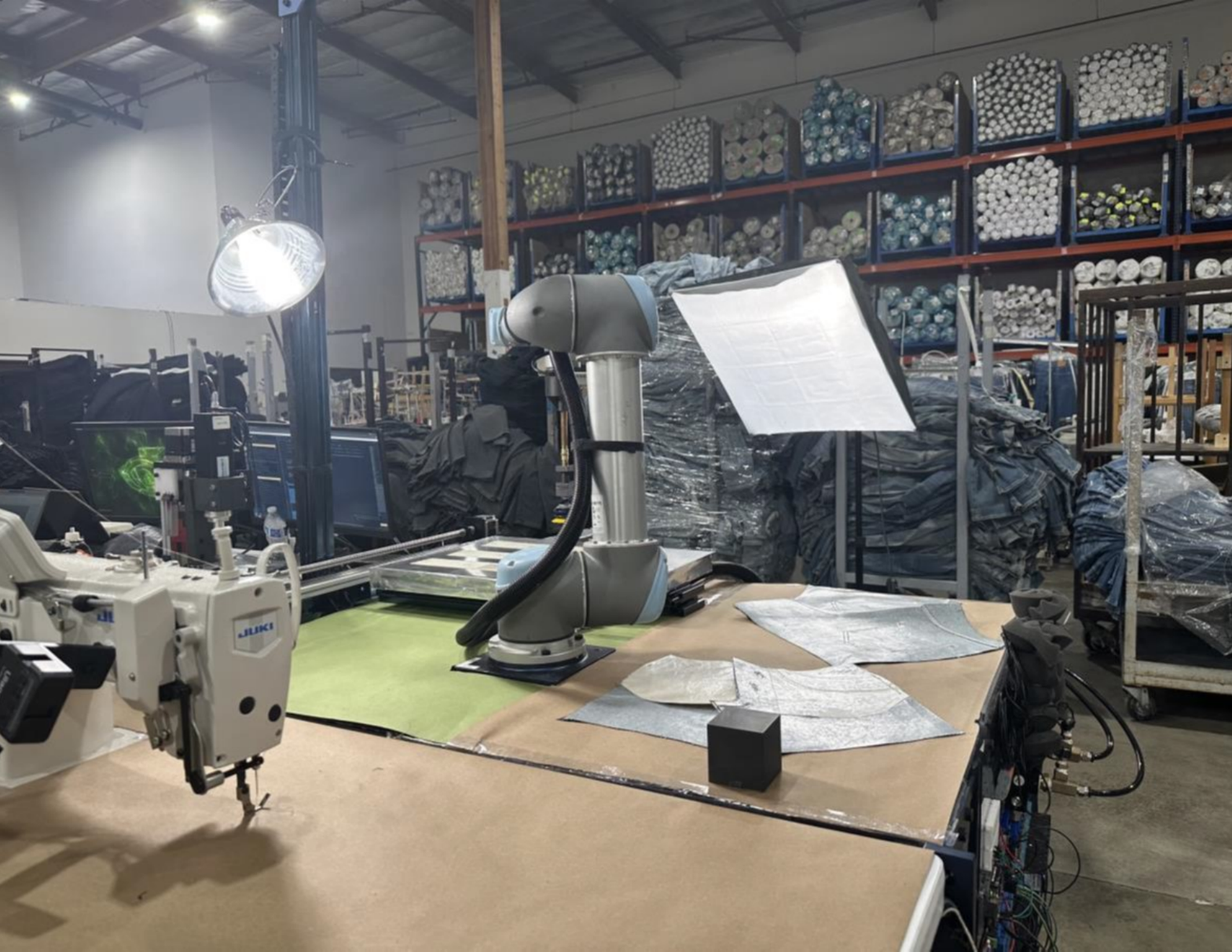}
  \caption{Robotic sewing workcell deployed on the factory floor for the staged denim automation.}
  \label{fig:workcell}
\end{figure}

However, these automated sewing systems rely on custom tooling, constrained hardware, or narrowly scoped operations. Most robotic garment assembly systems also remain focused on 2D sewing, while 3D sewing of curvilinear seams continues to limit broader adoption. Our recent work ~\cite{ajith2025roboticautomationapparelmanufacturing} introduced an end-to-end robotic sewing pipeline based on temporarily stiffened garment pieces, vacuum gripper based handling of the pieces, vision based piece alignment, temporary attachment, and closed-loop visual servoing, and demonstrated the feasibility of robotic sewing with conventional industrial equipment. Here, we build on that foundation from a deployment perspective. Through a denim manufacturing case study on the shop floor as shown in Figure~\ref{fig:workcell}, we examine the challenges and lessons learned during the development and deployment of such robotic sewing automation system, with emphasis on digital twin and digital thread integration, heterogeneous equipment interoperability, runtime verification, and operator-facing training.

The contributions of this paper are:
\begin{enumerate}
    \item A deployment case study of a robotic sewing workcell for denim manufacturing.
    \item Practical insights on digital twins, digital-thread integration, interoperability, runtime verification, and operator training.
\end{enumerate}

\section{Deployment Challenges in Apparel Automation}

\subsection{Material Handling}
Material variability remains a central challenge in apparel automation: even within a single garment category (e.g., denim shorts), parts can differ in stretch, thickness, dimensions and sensitivity to environmental conditions, which affects pickup, alignment, and sewing. Temporary stiffening as mentioned in our earlier work \cite{ajith2025roboticautomationapparelmanufacturing} and suction handling make robotic manipulation feasible, but during deployement the system should be able to operate with respect to the above mentioned variations caused by environmental conditions.

\subsection{Interoperability}
Conventional Machines on the apparel manufacturing floor doesn't provide standardized interfaces, yet a robotic sewing automation cell must coordinate multiple semi-automated machines, sensors, pneumatics, and the robot. Practical deployment therefore requires an interoperability layer that can interface with legacy machines, publish their state, and support coordinated execution.

\subsection{Production-Line Integration}
Process selection is a deployment decision: candidate operations must fit the surrounding production line operations, minimize disruption, and meet throughput and cycle-time constraints. This typically motivates the use of digital twins of the process and production plant to asses the right candidate process for automation.

\subsection{Human-Machine Interface}
Adoption also depends on usability. Operator-facing interfaces must support setup, supervision, and troubleshooting without requiring robotics expertise, making training and reconfiguration part of the deployment architecture.

\begin{figure}[t]
  \centering
  \includegraphics[width=\linewidth]{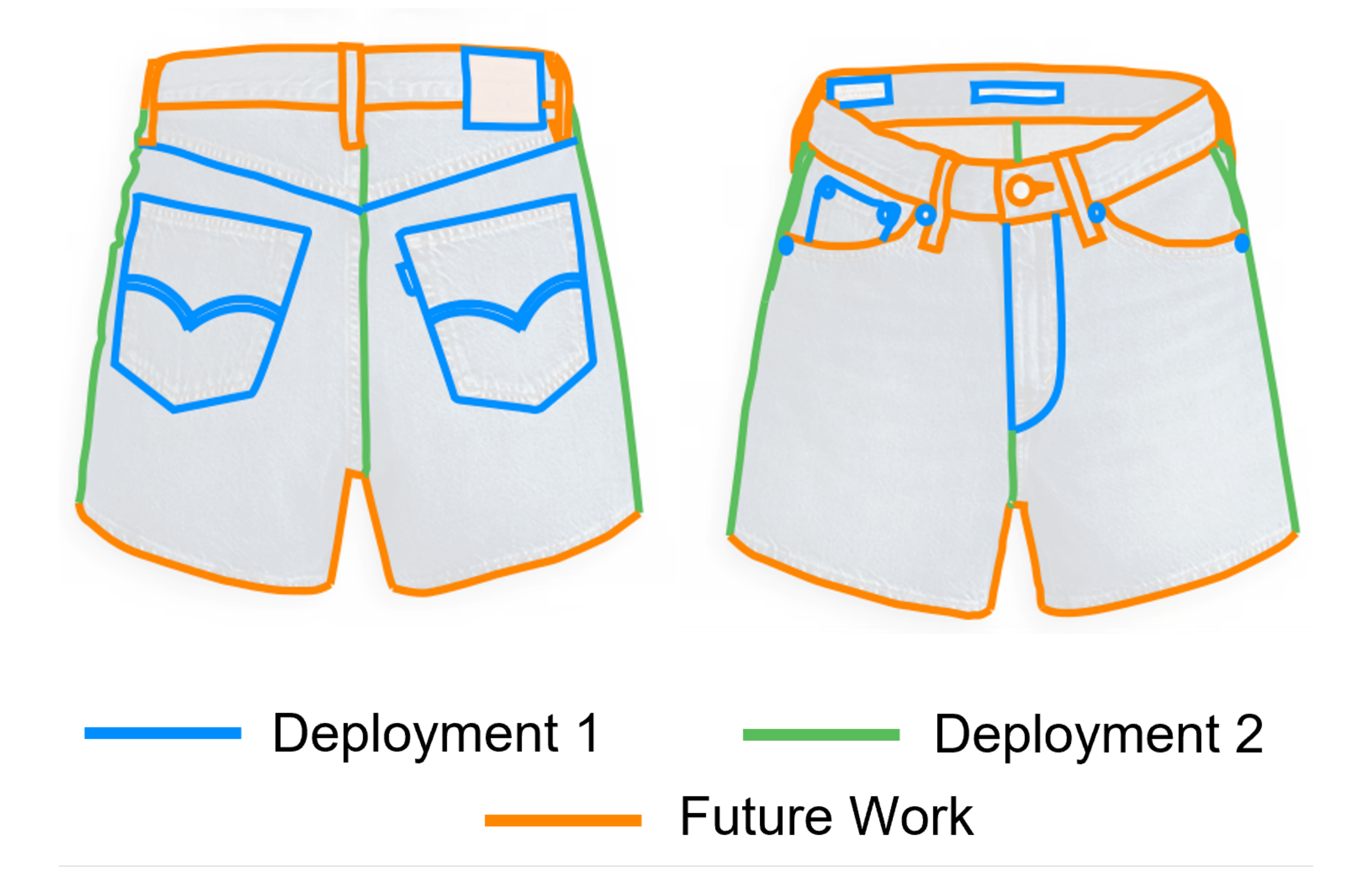}
  \caption{Denim shorts manufacturing design (back and front). Highlighted colors correspond to the operations targeted in the staged deployments.}
  \label{fig:denim-design}
\end{figure}

\section{Factory Deployment Case Study}
The deployments were executed on the factory floor in a production-adjacent parallel line (Fig.~\ref{fig:workcell}) to validate the system under realistic conditions without disrupting the primary production line.

\begin{figure}[t]
  \centering
  \includegraphics[width=\linewidth]{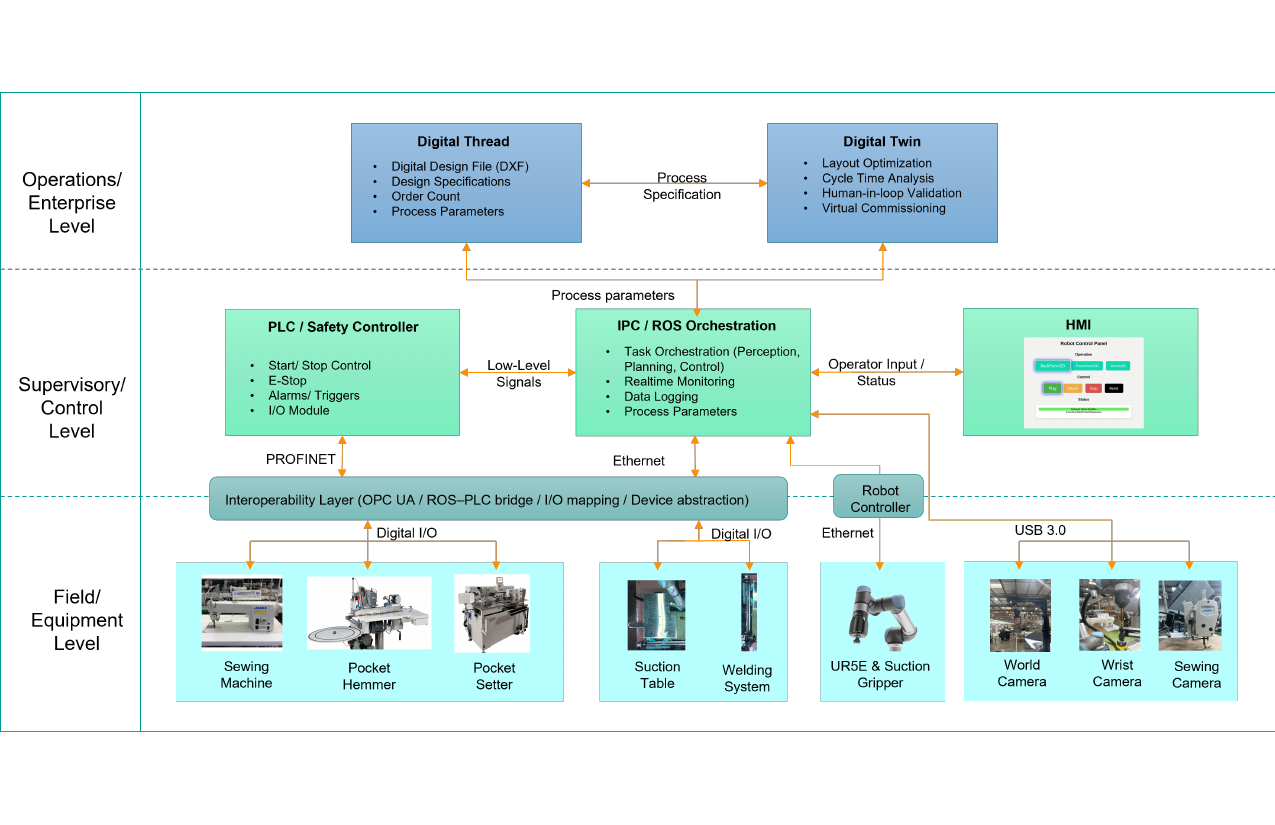}
  \caption{System architecture of the deployed robotic sewing workcell.}
  \label{fig:system_architecture}
\end{figure}

Our earlier work~\cite{ajith2025roboticautomationapparelmanufacturing} demonstrated that robotic automation with temporarily stiffened garments can succeed under controlled laboratory conditions. Moving toward commercial manufacturing, however, requires choosing processes and materials where stiffening can be integrated with minimal disruption and a credible return on investment. To guide this decision, we used digital models of the process and the plant in Siemens Plant Simulate and Process Simulate to compare candidate apparel workflows, evaluating material flow, cycle-time compatibility, and integration effort across process alternatives. This analysis identified sub-processes in denim shorts manufacturing as a practical first commercialization target. Crucially, the temporary stiffener must be removed after assembly, and denim production already includes a wash cycle that can remove, recover, and reuse the stiffener within the existing workflow, avoiding an additional downstream station. This compatibility lowers integration overhead and supports scalable deployment in a high-volume, relatively standardized product workflow.

To finalize the use case within denim manufacturing, we conducted a market study with Levi's to map the operation set involved in denim-shorts manufacturing and identify candidates for robotic automation. The study identified operations that could be automated by leveraging Sewbo's garment-stiffening technology, accounting for roughly 32\% of the overall operation set. Accordingly, Stage~1 focused on back-pocket assembly steps---pocket hemming, pocket setting, arcuate sewing, and tag placement (Fig.~\ref{fig:denim-design}, blue)---which accounted for 26\% of the overall operation set and provided an entry point for validating digital-twin integration, digital-thread execution, robotic sewing, and machine tending. Stage~2 extended the same workcell to shorts assembly, focusing on the front rise, back rise, and side seam operations that give rise to the 3D shape of the shorts (Fig.~\ref{fig:denim-design}, green).
% The overall system architecture is shown in Figure~\ref{fig:system_architecture}.
Although these 3D operations account for only about 6\% of the operation set, they are more challenging because they require coordinated 2D and 3D sewing to achieve the final garment shape.

The automation system uses a three-layer architecture, as shown in Fig.~\ref{fig:system_architecture}. At the operations/enterprise level, a digital thread extracts design specifications from DXF files, while a digital twin is used for layout optimization, cycle-time analysis, human-in-the-loop validation, and virtual commissioning. At the supervisory/control level, an industrial PC running the ROS-based orchestration layer coordinates perception, planning, control, runtime monitoring, and data logging, while a PLC handles start/stop logic, alarms, emergency-stop functions, and low-level I/O. An HMI provides operator interaction for status monitoring and start/stop. At the field/equipment level, the architecture integrates conventional sewing systems (including the sewing machine, pocket hemmer, and pocket setter), auxiliary process stations (e.g., the suction table and welding system), the UR5e robot with a suction gripper, and multiple cameras for world, wrist, and sewing views through an interoperability layer that standardizes interfaces for legacy systems. This layered architecture was essential for a successful factory-floor pilot deployment.

As summarized in Table~\ref{tab:deployment-metrics}, both deployment stages achieved 90\% trial success under factory conditions. We report their cycle times separately as mean $\pm$ std: Stage~1 (2D back-pocket subassembly) achieved $180\pm7$~s per trial, while Stage~2 (3D garment-shaping seams) achieved $360\pm12$~s per trial. In Stage~1, one trial corresponds to the full back-pocket subassembly sequence (pocket hemming, pocket setting, arcuate sewing, and tag placement), whereas in Stage~2 one trial corresponds to completing the front seam, side seam, and back seam. Success for each trial was assessed by manual quality inspection after completion: seam location was checked against a $1/4$~in tolerance, and alignment errors were required to remain within \SI{5}{\milli\metre} translational and \SI{2}{\degree} rotational tolerances; in future work, we will report a more detailed breakdown of the failure modes corresponding to the 10\% of failures.

\begin{table}[t]
  \centering
  \footnotesize
  \setlength{\tabcolsep}{4pt}
  \caption{Deployment trial outcomes and cycle times for the two staged factory deployments. Cycle times are reported as mean $\pm$ std (s).}
  \label{tab:deployment-metrics}
  \begin{tabular}{lccc}
    \toprule
    \textbf{Deployment} & \textbf{Trials} & \textbf{Success} & \textbf{Cycle time (s)} \\
    \midrule
    Stage 1 (2D) & 70 & 90\% & $180\pm7$ \\
    Stage 2 (3D) & 50 & 90\% & $360\pm12$ \\
    \bottomrule
  \end{tabular}
\end{table}

\section{Key Enablers for Deployment}\label{sec:key-enablers}

This section summarizes the key enablers that made the staged denim deployment practical on the shop floor.

\subsection{Digital Twin and Digital Thread Integration}
A key enabler for factory deployment was the combined use of a digital thread and a digital twin environment.
% (Fig.~\ref{fig:digital-thread}).
The digital thread provides a direct path from garment design specifications to executable robotic operations: the automation system processes the same digital drawings used in apparel production and automatically parses them into task parameters (e.g., seam geometry, feature locations, and target stitch paths) and calibration settings that are routed to the relevant subsystems. This matters on the factory floor because piece dimensions, tolerances, and even the operation sequence can vary across styles and lots; therefore, these specifications must be integrated into the deployment pipeline to reduce reprogramming effort.

The digital twin in Siemens Process Simulate complements this capability by making the tasks ``deployable'' in the context of the full workcell and surrounding workflow. In this work, the twin is used as an \emph{offline} commissioning and validation model (i.e., not a live, continuously synchronized digital twin during production). Before commissioning, we used the virtual workcell to (i) estimate cycle time and identify bottlenecks, (ii) test whether robotic operations could be synchronized with upstream and downstream stations, and (iii) iteratively refine the placement of major components (robot, fixtures, tables, and handoff points) to reduce transfers and nonproductive motion. Importantly, the twin was also used to evaluate the human-in-the-loop portions of the process, including operator approach paths, access zones for loading/unloading, and handoffs of intermediate parts to downstream processes. By verifying reach envelopes, task sequencing, and clearance zones in simulation, we reduced commissioning risk and limited disruption during on-floor integration.

Together, the digital thread and digital twin formed a practical deployment pipeline: the digital thread ensured that what the robot executed remained consistent with production design requirements and could be updated quickly, while the digital twin ensured that these tasks could be executed safely, meet cycle-time constraints, and remain compatible with operator workflows and the existing manufacturing line. \textbf{Takeaway: combining design-driven task generation with digital-twin validation reduces reprogramming effort and de-risks shop-floor integration.}

\subsection{Modular Interoperability and Workcell Orchestration}
Deployment also depended on an interoperability layer that could communicate with shop-floor equipment from different vendors---a major challenge in apparel automation, where production lines are dominated by legacy, stand-alone systems with limited or nonstandard digital interfaces. To introduce flexible automation without replacing existing systems, machine interfaces must be standardized through a communication layer that publishes machine state and accepts commands in a consistent way. Across the two denim deployments, the robot had to work with conventional sewing machines, semi-automated machines, welding stations, suction tables, and sensors. We therefore built the workcell around modular ROS components that communicate with low-level PLCs and microcontrollers through lightweight machine interfaces, enabling legacy equipment to be incorporated without major changes to the surrounding production line. This modularity also supported staged deployment, allowing simpler machine-tending operations to be validated first and then extended to more complex multi-station workflows. \textbf{Takeaway: interoperability is a key adoption bottleneck---a modular, vendor-agnostic interface layer lets brownfield machines report state and accept commands, enabling incremental automation without costly full line replacement.}

\subsection{Runtime Monitoring and Verification}
Our first deployment carried over an open-loop sewing strategy that had been validated using digital twins and controlled laboratory conditions. On the factory floor, however, environmental and upstream-process variability exposed failure modes that were not captured in initial testing: even within a single production lot, changes in humidity affected material properties, leading to curling and deformation of the workpiece that affected grasping and sewing operations. To address this in the second deployment, we introduced runtime monitoring and verification, including adaptive perception for grasping, vision-based alignment, and closed-loop sewing with visual servoing. We also integrated machine-level coordination and collision checking to ensure safe execution in the constrained workcell geometry. Together, these capabilities improved seam accuracy and acted as a verification layer that increased robustness to part variation, environmental conditions, and shop-floor variability. \textbf{Takeaway: shop-floor process variance cannot be fully accounted for during development and offline digital-twin testing; runtime verification with online correction is necessary to detect and recover from deviations during execution.}

\subsection{Operator-Facing Training and Reconfiguration}
Operator-facing support was an important deployment enabler, especially in apparel manufacturing where workers are familiar with sewing equipment but not necessarily with robotic systems. To address this, we developed device-agnostic digital training and guidance that could be accessed through phones, tablets, or head-mounted displays, with content shaped in collaboration with ISAIC to reflect workforce-upskilling needs. The materials included digital work instructions, step-by-step guidance, QR-based access near the workcell, and troubleshooting support for setup and reconfiguration. More broadly, the same digital twin environment used for system design and commissioning can also support operator training by allowing users to understand workflows, cell behavior, and troubleshooting procedures in a safe virtual setting before interacting with the physical workcell. While operator training was not the primary focus of this work, these needs became clear primarily during deployment rather than during development. \textbf{Takeaway: incorporate operator guidance and training during development (ideally using the digital twin) to surface training issues early and reduce deployment ramp-up risk.}

\section{Conclusion}

This paper presented a deployment-oriented case study of a robotic apparel automation system for denim manufacturing that integrates digital thread with a digital twin pipeline to move from process validation to factory-floor execution. Across two staged deployments, we addressed key deployment barriers: material handling, interoperability with heterogeneous legacy equipment, and shop-floor variability through a modular interoperability communication layer and runtime monitoring and correction. The resulting system architecture and deployment lessons provide a practical blueprint for scaling such systems beyond lab demonstrations.

%%%%%%%%%%%%%%%%%%%%%%%%%%%%%%%%%%%%%%%%%%%%%%%%%%%%%%%%%%%%%%%%%%%%%%%%%%%%%%%%

\section {ACKNOWLEDGMENT}
The Research was sponsored by the ARM (Advanced Robotics for Manufacturing) Institute through a grant from the Office of the Secretary of Defense and was accomplished under Agreement Number W911NF-17-3-0004. The views and conclusions contained in this document are those of the authors and should not be interpreted as representing the official policies, either expressed or implied, of the Office of the Secretary of Defense or the U.S. Government. The U.S. Government is authorized to reproduce and distribute reprints for Government purposes notwithstanding any copyright notation herein.

%%%%%%%%%%%%%%%%%%%%%%%%%%%%%%%%%%%%%%%%%%%%%%%%%%%%%%%%%%%%%%%%%%%%%%%%%%%%%%%%

\addtolength{\textheight}{-12cm}   % This command serves to balance the column lengths

%%%%%%%%%%%%%%%%%%%%%%%%%%%%%%%%%%%%%%%%%%%%%%%%%%%%%%%%%%%%%%%%%%%%%%%%%%%%%%%%

%%%%%%%%%%%%%%%%%%%%%%%%%%%%%%%%%%%%%%%%%%%%%%%%%%%%%%%%%%%%%%%%%%%%%%%%%%%%%%%%

%%%%%%%%%%%%%%%%%%%%%%%%%%%%%%%%%%%%%%%%%%%%%%%%%%%%%%%%%%%%%%%%%%%%%%%%%%%%%%%%

% Bibliography (BibTeX)

% \bibitem{robotassembly2020}
% \url{https://arminstitute.org/projects/robotic-assembly-of-garments/}

% \bibitem{plantsimulate}
% \url{https://www.siemens.com/en-us/products/tecnomatix/plant-simulation-software/}

% \bibitem{processsimulate}
% \url{https://www.siemens.com/en-us/products/tecnomatix/process-simulate-software/}
\bibliographystyle{IEEEtran}
\bibliography{ref}

\end{document}